\documentclass[conference]{IEEEtran}
\IEEEoverridecommandlockouts
\usepackage{cite}
\usepackage{amsmath,amssymb,amsfonts}
\usepackage{mathtools}
\usepackage{algorithmic}
\usepackage{graphicx}
\usepackage{textcomp}
\usepackage{xcolor}
\usepackage{multirow}
\usepackage{booktabs}
\usepackage{subcaption}
\usepackage{url}
\usepackage{hyperref}
\usepackage[font={footnotesize}]{caption}
\def\BibTeX{{\rm B\kern-.05em{\sc i\kern-.025em b}\kern-.08em
    T\kern-.1667em\lower.7ex\hbox{E}\kern-.125emX}}

\begin{document}

\title{Robot Air Hockey: A Manipulation Testbed for Robot Learning with Reinforcement Learning\\
\thanks{* denotes equal contribution. For corresponding author and code requests contact \texttt{calebc@cs.utexas.edu}}
}

\author{\noindent 
Caleb Chuck\(^{1,*}\), 
Carl Qi\(^{1,*}\), 
Michael J. Munje\(^{1,*}\), 
Shuozhe Li\(^{1,*}\), 
Max Rudolph\(^{1,*}\), \\ 
Chang Shi\(^{1,*}\), 
Siddhant Agarwal\(^{1,*}\), 
Harshit Sikchi\(^{1,*}\), 
Abhinav Peri\(^{1}\), Sarthak Dayal\(^{1}\),\\ Evan Kuo\(^{1}\), Kavan Mehta\(^{1}\), Anthony Wang\(^{1}\), Peter Stone\(^{1,3}\), Amy Zhang\(^{1}\), Scott Niekum\(^{2}\) \\ 
\\ \(^{1}\) The University of Texas at Austin \\ $^2$ University of Massachusetts Amherst \\ $^3$ Sony AI}

\maketitle
\begin{abstract}
Reinforcement Learning is a promising tool for learning complex policies even in fast-moving and object-interactive domains where human teleoperation or hard-coded policies might fail. To effectively reflect this challenging category of tasks, we introduce a dynamic, interactive RL testbed based on robot air hockey. By augmenting air hockey with a large family of tasks ranging from easy tasks like reaching, to challenging ones like pushing a block by hitting it with a puck, as well as goal-based and human-interactive tasks, our testbed allows a varied assessment of RL capabilities. The robot air hockey testbed also supports sim-to-real transfer with three domains: two simulators of increasing fidelity and a real robot system. Using a dataset of demonstration data gathered through two teleoperation systems: a virtualized control environment, and human shadowing, we assess the testbed with behavior cloning, offline RL, and RL from scratch. 
\end{abstract}


\begin{figure*}[t]
    \centering
    \begin{subfigure}[t]{0.62\linewidth}
    \centering
    \includegraphics[width=\linewidth]{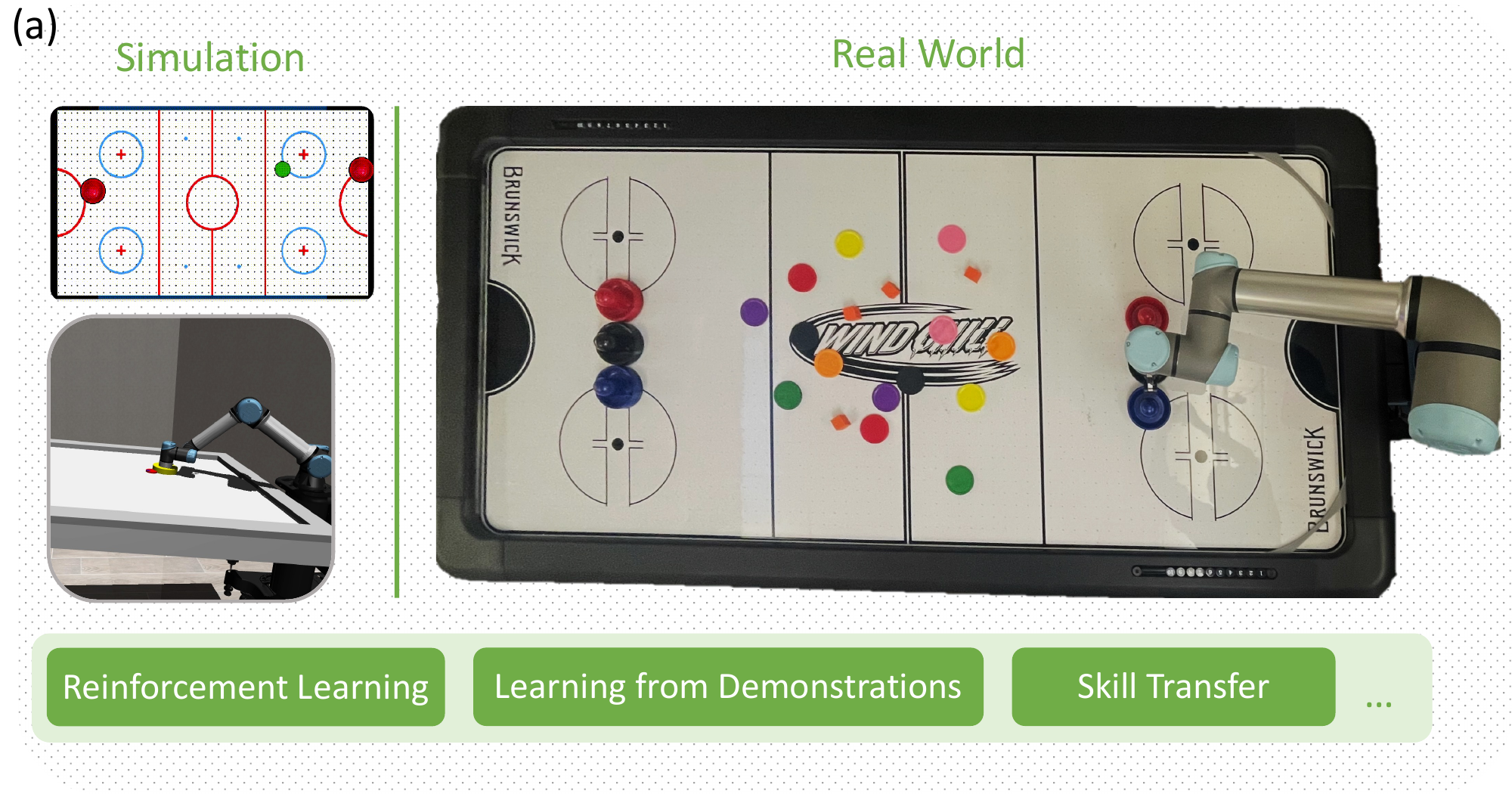}
    \end{subfigure}
    ~
    \begin{subfigure}[t]{0.32\linewidth}
    \centering
    \includegraphics[width=\linewidth]{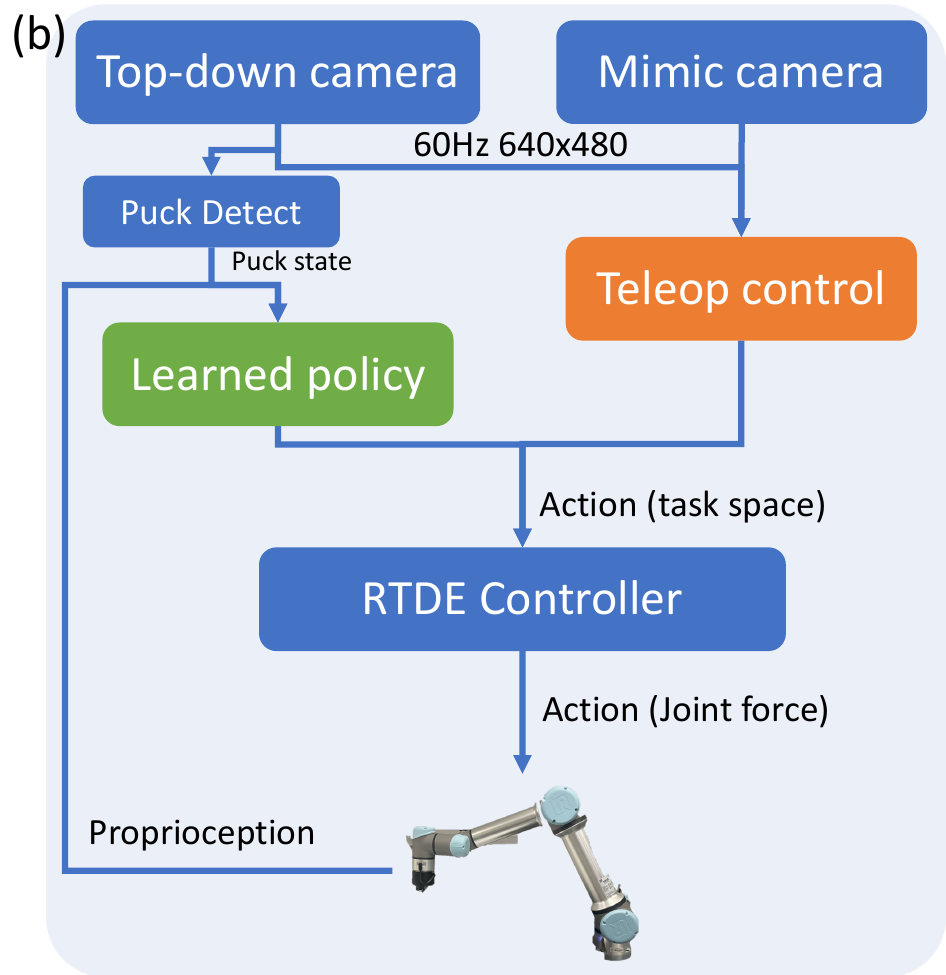}
    \end{subfigure}
    \caption{(a) Robot Air Hockey is a testbed that contains a large number of dynamic, interactive Air Hockey tasks in multiple distinct simulators as well as the real world. It is suitable for evaluating a variety of frameworks from vanilla RL (both online and offline), learning from demonstrations, to skill transfer or goal-conditioned RL. (b) Overview of our control pipeline. We use object detection to get the state of interactive objects, and the UR5 RTDE controller to transform the task actions into joint forces for the robot.}
    \label{fig:pull}
\end{figure*}
\section{Introduction}
Reinforcement Learning (RL) offers a promising direction for real-world robotics by allowing robotics to accomplish complex tasks using only a reward description.
Although high-performance demonstrations can sometimes be collected through tools like teleoperation, it is not always feasible (and often not possible) to collect large quantities of such ``expert'' data.
Recent developments in RL~\cite{schrittwieser2020mastering,brohan2023rt,team2023octo} offer ever-increasing generalization capabilities, and improved learning algorithms through goal-conditioned RL~\cite{chebotar2021actionable,sikchi2023score,eysenbach2021replacing} or unsupervised skill learning~\cite{eysenbach2018diversity,park2023metra,zahavy2022discovering,chuck2023granger,chuck2020hypothesis}, as well as utilization of human information such as preferences and offline demonstrations~\cite{ahn2022can,hejna2023contrastive,sikchi2022ranking,sikchi2023dual,ni2021f}. Suitable testbeds for assessing RL in the real world will help scale RL to real-world robotics.

Real-world environments are challenging because they are often both dynamic and interactive. For example, objects can roll about the table when cooking or might fall when cleaning. The world is full of \textit{dynamic elements}---state features that are constantly moving. By contrast, \textit{quasistatic elements} remain predominantly stationary and only move when being operated on by the robot. Furthermore, real-world tasks often require the agent to \textit{interact} with its environment, i.e. making contact with and even manipulating, elements such as objects and other agents. In tasks that are both dynamic and interactive, RL offers a promising direction since human demonstrators can often struggle with precise, high-speed robot teleoperation, and hard-coded policies can be brittle when taken out of a controlled context.

To evaluate RL in these settings, a testbed should support the complexity of the real world while also allowing assessment of the variety of RL settings. Reinforcement Learning research has introduced a wide gamut of tools, such as goal-conditioning, model learning, task transfer, skill learning, offline, and inverse reinforcement learning---to name just a few. While many of these tools have been applied to simulated environments, even simulated robots, testing these methods on a physical robot is challenging. Offering a single testbed that supports these settings both in simulation and with a real-world system would be an important step for the improved evaluation of RL in robotic manipulation.

\begin{figure}[ht]
    \centering
    \includegraphics[width=0.6\linewidth]{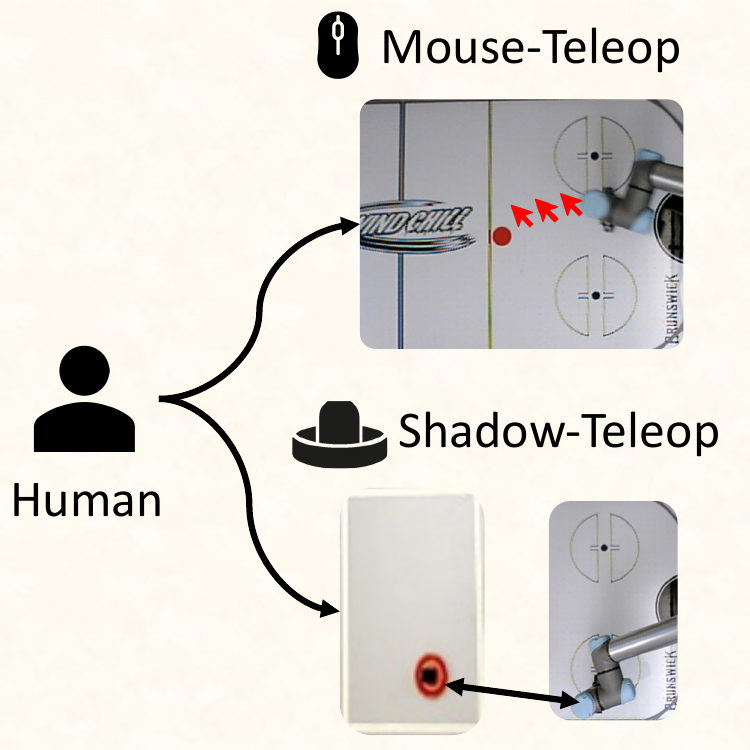}
    \caption{Robot Air Hockey supports two types of teleoperations. Mouse-Teleop (top): The user moves the mouse to control the robot. Shadow-Teleop (bottom): the user moves a paddle to control the robot.}
    \label{fig:teleop}
\end{figure}

We introduce a novel dynamic, interactive RL testbed that modifies air hockey, a popular game, with a collection of objects, Figure~\ref{fig:pull} illustrates some of the potential of this domain. By focusing on puck-hitting, the domain is inherently interactive and dynamic. 
This platform offers several advantages that facilitate RL training. The puck's constrained movement allows for efficient environment resets, while a strictly controlled agent workspace ensures safe autonomous operations when exploring. 
By incorporating multiple objects, both virtual and real, we can describe a wide array of tasks, illustrated in Figure~\ref{fig:realworld_tasks}, allowing for the assessment of goal-conditioned, transfer, or skill learning methods. To evaluate RL without the physical setup we provide \textit{two} simulators, illustrated in Figure~\ref{fig:pull}, of increasing fidelity to the real world and tunable parameters so that sim-to-real transfer algorithms can be assessed even without the physical system, by proxy of sim-to-sim transfer. Furthermore, we introduce a human-teleoperated dataset using two teleoperation systems for the agent which can be used in the simulators and the real world, allowing the assessment of learning from demonstration and offline RL algorithms (see Figure~\ref{fig:teleop}).

This work not only describes the domain and associated tasks but also assesses several baseline algorithms and elucidates the key design decisions motivating the algorithm. We also assess behavior cloning, vanilla RL, and offline RL on several tasks in simulation and the real world. We demonstrate empirically that the set of tasks developed has a smooth variation from easy to difficult for all of these algorithms, both in simulation and the real world. Together, this testbed provides a dynamic, interactive environment where existing algorithms tested with other benchmarks could struggle. Algorithms assessed with quasistatic elements that rely on the slow movement of the robot could be exposed to the fast-moving dynamics, and those without interactive elements could struggle to adapt to the sparse puck or object interactions. This testbed offers a tool for assessing a wide range of RL algorithms on a similar footing in a real-world setting.


\begin{table*}[t]
    \centering
    \scalebox{0.9}{
    \begin{tabular}{cc|cccccccccccc|}
    \toprule
    Environment & Method &
     
     \multicolumn{10}{c}{Robot Air Hockey Tasks} \\
     \midrule
      & & Reach & Reach V. & Touch & Strike & Strike Crowd & Juggle & Puck V. & Block & Hit Goal & Hit Goal V. \\ 
    \midrule
    \multirow{4}{*}{Box2D~\cite{Box2D}} 
    & BC & 0.9  & 0.8 & \textbf{1.0} & 0.7 & 0.3 & 0.3 & 0.7 & 0.0 & 0.1 & 0.0 \\
    & IQL & \textbf{1.0} & \textbf{1.0} & \textbf{1.0} & \textbf{1.0} & \textbf{1.0}& \textbf{1.0} & \textbf{1.0} & 0.0 & 0.4 & 0.0 \\
    & RL & \textbf{1.0} & \textbf{1.0} & \textbf{1.0} & \textbf{1.0} & \textbf{1.0} & \textbf{1.0} & \textbf{1.0} & 0.0 & \textbf{0.9} & 0.0\\
    \midrule
   
    \multirow{4}{*}{Robosuite~\cite{zhu2020robosuite}} 
    & BC & 0.9 & 0.8 & 0.8 & - & - & 0.6 & 0.6 & - & 0.1 & - \\
    & IQL & 0.9 & 0.9 & 0.8 & - & - & 0.7 & 0.8 &- & 0.1 & - \\
    & RL & \textbf{1.0} & \textbf{1.0} & \textbf{1.0} & - & - & \textbf{0.9} & \textbf{0.9} & - & \textbf{0.2} & -\\
    
    \midrule
    \multirow{4}{*}{Real World} & BC & 0.9 & \textbf{0.1} & 0.3 & - & - & - & 0.1 & - & - & - \\
    & IQL & \textbf{1.0} & 0.0 & 0.6 & - & - & - & 0.3 & - & - & - \\
    & Human & \textbf{1.0} & 0.0 & \textbf{1.0} & - & - & \textbf{0.3} & \textbf{1.0} & - & - & - \\

    \bottomrule
    \end{tabular}}
    \caption{Quantitative results for the Box2D, robosuite and real robot environment. A dash (-) indicates that this combination of environment, method and task was not evaluated. Simulation performances are averaged across 5 seeds.
    For Box2D, a combination of SAC/HER \cite{haarnoja2018soft, andrychowicz2017hindsight} is used to train the goal-conditioned RL policies, while PPO \cite{schulman2017proximal} is used to train RL policies for the remaining tasks. For Robosuite, we use PPO for Vanilla RL, following the implementation of CleanRL library \cite{huang2022cleanrl}. 
    In simulation, BC and IQL \cite{kostrikov2021iql} are used to learn policies offline using the ``expert'' data collected with trained PPO policies. On the real robot, we instead use a teleoperated dataset collected by 8 human players of varying skill of 400 trajectories for BC and IQL.
    }
    \vspace{-5mm}
    \label{tab:exp-main}
\end{table*}

\section{Related Works}
\label{sec:related}
Agile robots and tasks, distinguished by fast, dynamic motion, have seen growing interest in the past several years. One instantiation of an agile robotic task is autonomous air hockey and has been studied in several settings~\cite{jan_air_hockey, taitler2017learning, jankowski2024airlihockey,airhockey1,airhockey2,airhockey4,airhockey5,airhockey6,airhockey7,airhockey8,airhockey9}. Notably, a recent workshop~\cite{robotair} highlighted robot air hockey as a competitive robot learning task with several submissions on model-based and deep learning methods \cite{orsulalearning, deair, minnucci2022applying}. Specifically, Liu et al.\cite{jan_air_hockey} used Bayesian optimization and a slew of other well-tuned solvers to plan trajectories for striking the puck. While this hand-crafted technique is successful, it is difficult to generalize across small variations in air hockey table physical parameters and task specifications. Most autonomous robotic air hockey  systems focus on learning to play competitive matches with humans. In contrast, our work focuses on a \textit{variety} of tasks, from goal conditioned puck hitting to puck juggling. Furhter, we incorporate of human teleoperation modes to assess learning from demonstration and other offline methods. The many works on robotic air hockey cover a wide range of topics and are summarized here:
\begin{enumerate}
    \item Namiki et al~\cite{airhockey1} uses a hierarchical switching controller to play with different behaviors according to the robot's opponent.
    \item Owgawa et al~\cite{airhockey2} aims to generate motion plans that take advantage of weak spots in a human's vision when switching gaze.
    \item Bishop et al.~\cite{airhockey4} demonstrates the difficulty of building a vision-based autonomous air-hockey system.
    \item AlAttar et al.~\cite{airhockey5} presents a state-model-based method to create air hockey-playing agents against which humans can compete.
    \item Alizadeh et al.~\cite{airhockey6} introduces a method to estimate the parameters of air hockey tables (e.g. coefficient of friction, etc.) automatically so that a robot may adapt its behavior accordingly.
    \item Liu et al.~\cite{airhockey7} demonstrates a fast-hitting, air hockey-playing robot using analytical controllers that were developed by modeling the air hockey environment. 
    \item Tadokoro et al.~\cite{airhockey8} design a high-speed wrist mechanism as an add-on to a manipulator so that a robot can easily compete with the speed of a human air hockey player. 
    \item Huang et al.~\cite{airhockey9} introduces a method to align simulators with real-world observations using causal relationships.
\end{enumerate}
Robotic air hockey is an active field of research with a diverse set of reliable solutions, most notably reinforcement learning. Our testbed is well-positioned to act as a benchmark for learning agile manipulation skills.

\section{Domain Description}

\subsection{Components}
The common components across environments consist of the table, the paddle, the pucks, moveable objects such as blocks, and immovable obstacles. A robot arm is used in the Robosuite and real-world environments to control the paddle. In the 2D simulator the paddle is manipulated directly. In all domains, the table workspace is $66 $ in $ \times 24 $ in, the paddle is $3.75$in in diameter and the puck is $2.5$ in. We maintain consistency between the domains to leave open the possibility of transfer. 

\subsection{Teleoperation}
\subsubsection{Mouse-Teleop}
The mouse teleoperation setup streams a live video feed of the robot as seen by the overhead camera after performing an orthographic tomography to transform the camera coordinate system into the robot coordinate system in 2D. The mouse $x,y$ position in the image is mapped to a desired robot $x,y$ pose, using force control to maintain contact with the table. Desired positions are then projected into the bounded action distances. This simple approach to teleoperation allows us to easily collect data for various tasks.

\begin{figure}[ht]
    \centering
    \includegraphics[width=0.8\linewidth]{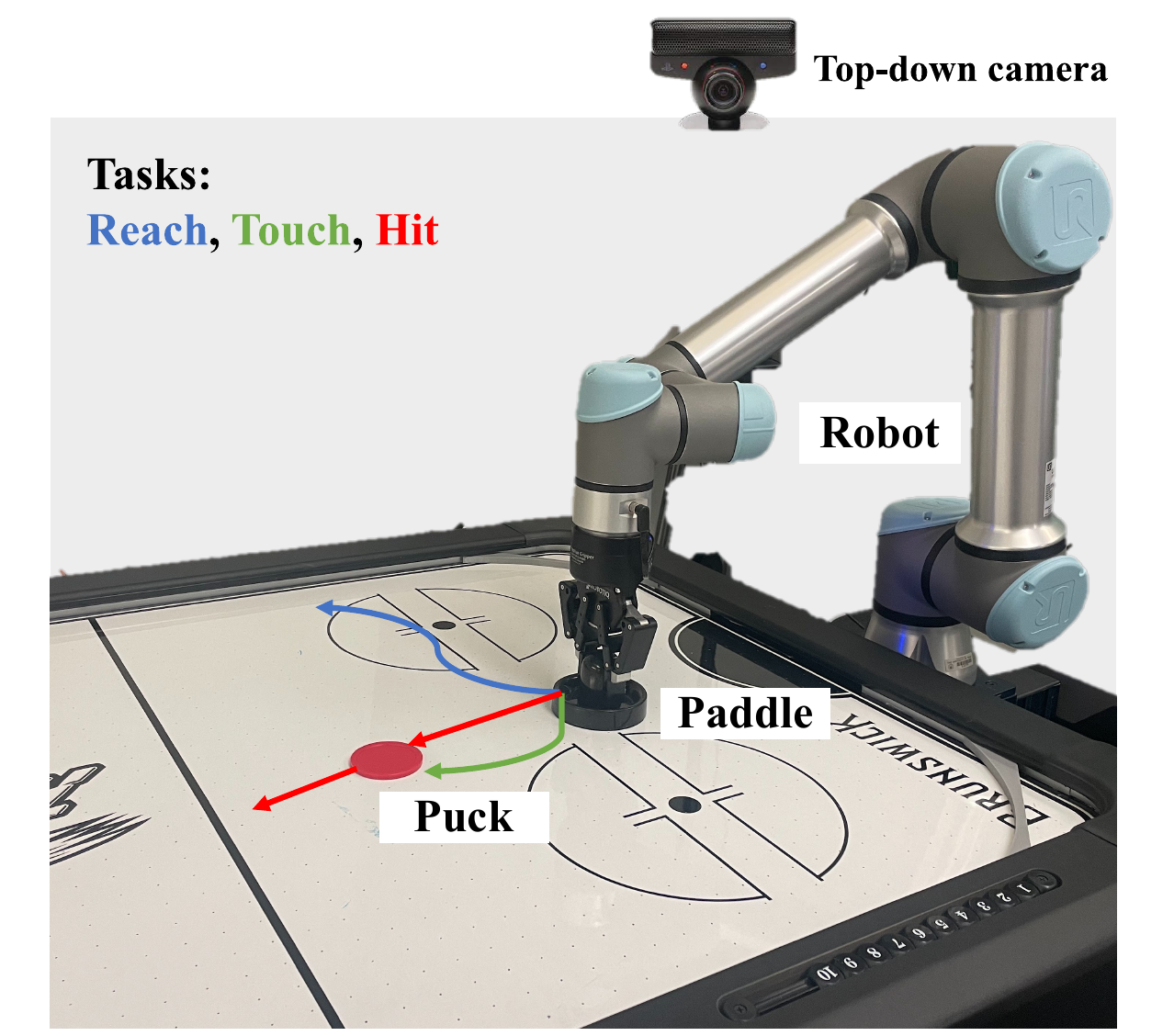}
    \caption{Robot Air Hockey real-world setup. We use a top-down camera to provide observation and a UR5e robot to actuate the paddle. Our real-world setup can facilitate many air hockey tasks, including but not limited to reaching, touching, and hitting.}
    \label{fig:realworld_tasks}
    \vspace{-2mm}
\end{figure}

\subsubsection{Shadow-Teleop}
The shadowing setup consists of a flat surface, a paddle, and an above camera. We utilize color segmentation to track the paddle movement on this surface and transform these movements to end-effector locations on the robot. This approach allows the user to play more naturally, as it is in direct control of a paddle, while bounded by the capabilities of the robot.

\subsection{Environments}

A common high-level interface is used for all environments, allowing the same reward functions to be used across all individual environments. All environments share a common state and action space. 

\subsubsection{2D}
Our 2D simulation environment uses the Python implementation of Box2D \cite{Box2D} as a physics simulation backend. While we are not able to deploy robot arm models in this environment, we use hand-tuned reward shaping so that sequences of actions are empirically more realistic for a robot arm. This environment has many changeable world parameters such as paddle mass, puck mass, dampening, friction, gravity, starting puck velocities, and many more parameters. 
\subsubsection{3D}
Our 3D simulation environment is a custom Robosuite \cite{zhu2020robosuite} setting which builds on top of a MuJoCo \cite{todorov2012mujoco} as a simulation backend and allows us to simulate the robot arm in addition to the other components such as the table, a paddle, and a puck. For control, the operational space controller \cite{osc} is available in Robosuite, which we modify to maintain stable contact with the table.

\subsubsection{Real World}

Our real-world setup consists of a 76in x 34in Wind Chill air hockey table tilted $5.5$ degrees and a UR5e robot arm. The vision system uses a Sony Playstation Eye, a high framerate camera, which gathers $640\times480$ frames at $60$ FPS, mounted to the ceiling to have a full view of the table. The robot control operates at $20$ FPS, which is dictated by the cost of vision processing and computing the desired action from the model while maintaining a stable control loop. We use hue-saturation-value (HSV) segmentation and a homography transformation using OpenCV \cite{opencv_library} to estimate puck location and velocity. We visualize puck detection in Figure~\ref{fig:puck_detect}. We use force and operation space control on the robot to maintain the contact between the paddle and the table. A picture of our workspace and tasks is described in Fig~\ref{fig:realworld_tasks}. The control flow is illustrated in Figure~\ref{fig:pull}b.

\subsection{Tasks}
We provide a collection of ten total tasks which vary in difficulty from those that can be achieved through behavior cloning (reaching) to those where even humans can struggle in the real world (juggling). Each task includes a designed reward function, and some can include additional objects beyond the puck and the end effector, such as target regions or other objects to be manipulated through the puck. We have \textbf{five tasks} on the real robot, \textbf{six tasks} in Robosuite, and \textbf{ten tasks} in Box2D. The wide variety of tasks allows us not only to assess varying difficulty but also leaves room for potential skill learning and transfer learning. 

Below we describe the ten tasks we assessed in this work. In addition to the reward specified by the task itself, we also provided regularization to ensure that undesirable behaviors such as jittering and twisting motion, which can cause the robot to emergency stop (a safety measure on the UR5 that prevents it from damaging itself or the human), were prevented.
\begin{enumerate}
\item \textbf{Reaching}: The paddle reaches a random location. When the paddle is within $\epsilon$ of the goal, the agent receives a positive reward and the episode is reset.
\item \textbf{Reaching with Velocity}: The paddle reaches a random location \textit{and velocity} combination. When the paddle is within $\epsilon_\text{position}, \epsilon_\text{velocity}$ of the goal (both position and velocity) a reward is given, and the episode is reset.
\item \textbf{Touching}: The paddle touches the puck. Upon detection of contact, a reward is given. The agent is continually rewarded each time it touches the puck.
\item \textbf{Striking a stationary puck}: The paddle hits a stationary puck and moves it a minimum distance. If a sufficient velocity is achieved, a reward is given and the episode ends. 
\item \textbf{Striking a stationary puck into a crowd}: The paddle hits a stationary puck which causes it to collide with blocks, similar to the game of pool. The reward is dependent on the amount of spread from the blocks from the crowd.
\item \textbf{Juggling}: The paddle hits a puck a minimum distance above the paddle at the time of hitting.
\item \textbf{Puck Velocity}: The paddle hits a puck and causes it to move at a minimum upward velocity.
\item \textbf{Moving a block}: The paddles hits a puck into a block, causing it to move a minimum distance from the block's initial position.
\item \textbf{Hitting into a goal region}: The paddle hits a puck into a goal circle region with a constant radius.
\item \textbf{Hitting into a goal region with desired velocity}: The paddle hits a puck into a goal circle region with a constant radius with a specified velocity. It is rewarded based on distance to the goal region's center, cosine similarity between the puck's vector when entering the goal region, and difference in the puck's magnitude from the desired velocity's magnitude.
\end{enumerate}

While not a task we trained models for in this work, our Box2D simulation environment also supports both collaborative play and adversarial play, extending these environments as a potential testbed for multi-agent RL. Furthermore, additional tasks with increasing complexity can be easily constructed.


\subsection{Offline Data}
\begin{figure*}[ht]
    \centering
    \centering
    \includegraphics[width=0.8\linewidth]{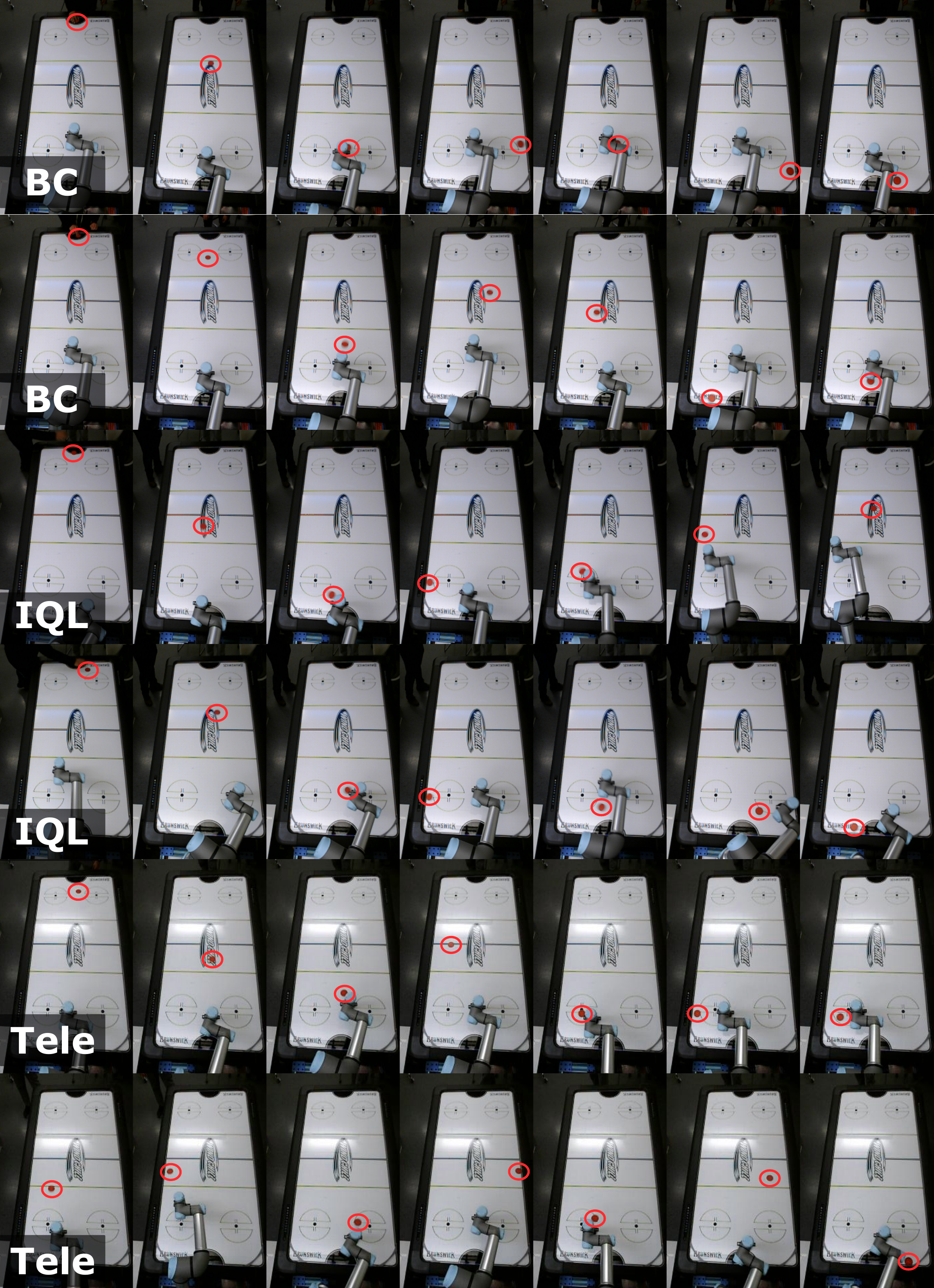}
    \caption{Execution rollouts on the UR5 air hockey setup for policies trained with behavior cloning, IQL with the touching loss, and human teleoperation demonstrations. Notice that even though humans are trying to achieve multiple bounces, they often hit the puck too erratically to effectively return, so demonstrations can vary significantly in skill, even after cleaning human failure modes.}
    \label{fig:rollouts_real}
\end{figure*}

On the real robot, we provide a dataset of human gameplay using both teleoperation methods ($350$ mouse-teleop trajectories and $50$ shadow-teleop trajectories) gathered from 8 participants of varying skill. We visualize some of the human-gathered demonstrations in Figure~\ref{fig:rollouts_real} and Figure~\ref{fig:puck_detect}. In these, it is clear that the human demonstrators, while able to hit the puck at least once, can struggle to achieve multiple hits. In Table~\ref{tab:exp-main}, we assess successful juggling as hitting the puck at least four times in a trajectory, and we can see that humans only achieve moderate success. We discuss real-world data collection and visualize human-gathered policies in Figure~\ref{fig:rollouts_real} and Appendix~\ref{sec:puck_detection}.

\section{Learning Methods}

In our experiments, we evaluate three representative methods: Behavior cloning (mean squared error loss), Vanilla Reinforcement Learning (SAC~\cite{haarnoja2018soft} and PPO~\cite{schulman2017proximal}), and Offline Reinforcement Learning (IQL~\cite{kostrikov2021iql}). Behavior cloning learns a policy to take desired actions, Reinforcement learning maximizes discounted reward, and offline RL assumes environment interactions are not available to the agent.

\subsection{Preliminaries}
A Markov decision process is defined by the tuple $\mathcal M \coloneqq (\mathcal S, \mathcal A, p, R)$, where $\mathcal S$ is the state space, $\mathcal A$ is the action space and $s\in \mathcal S, a \in \mathcal A$ are states and actions respectively. $p(s'|s,a)$ is the transition function that gives the probability of the next state $s'$ given the current state and action $(s,a)$. The reward function $R(s,a)$ maps state and action to a scalar reward. A policy $\pi(a|s)$ is the probability of an action given the current state.

\subsection{Behavior Cloning}
Behavior cloning~\cite{pomerleau1991efficient} is an offline imitation learning approach that learns a policy using supervised learning. We use the common MSE loss to learn $\pi$:

$$\mathcal{L}_\text{bc} = \mathbb{E}_{(a,s) \sim \mathcal{D}}[\| \pi(s) - a \|^2_2].$$

Although the performance of behavior cloning is limited by the quality of the dataset and is plagued by compounding error problems in the limited data regime, it presents a simple and scalable alternative~\cite{ahn2022can,jang2022bc} to the more complicated yet unstable imitation learning methods~\cite{ho2016generative,sikchi2022ranking,ghasemipour2020divergence}.

\subsection{Reinforcement Learning}

The goal of Reinforcement Learning is to learn a policy maximizing returns obtained on any user-specified reward function. Formally, RL aims to learn a policy $\pi$ that maximizes the cumulative return $J(\pi) = \mathbb{E}_{\pi}[\sum_{t=0}^{\infty} \gamma_t r(s_t,a_t))]$. RL reduces the effort of designing hand-tuned controllers by proposing general approaches that learn any behaviors. 

\textbf{Online RL}: In our work, we rely on Proximal Policy Optimization (PPO)~\cite{schulman2017proximal} as our baseline algorithm. PPO has been extensively used in robotics for its stability and guarantee of near-monotonic improvement.

\textbf{Offline RL}: Offline RL deals with a specific setting where environment interactions are not available to the agent, but rather the agent is provided with data of offline transitions of the form \{s,a,r,s'\} from the environment. This setting reduces the burden of exploration considerably and provides a safe way to learn from previously collected datasets. Offline deep RL algorithms suffer optimization difficulties due to problems like overestimation~\cite{fujimoto2018off}, and feature-coadaptation~\cite{kumar2021dr3} leading to popular regularizers like pessimism~\cite{fujimoto2021minimalist,levine2020offline,sikchi2023dual}. We use a representative and performant algorithms for offline RL in this work - IQL~\cite{kostrikov2021iql}. IQL is a modification on the standard actor-critic learning procedure to replace the maximization of the state-action value function with a maximization that only chooses values for actions seen in the dataset.
\\ \\ 
For all of our experiments, our policies are represented with multi-layer perceptions with a hidden layer of 256. Further training details are described in Appendix~\ref{sec:training_curves}.

\section{Experimental Results}

\begin{figure*}[ht]
    \centering
    \centering
    \includegraphics[width=0.8\linewidth]{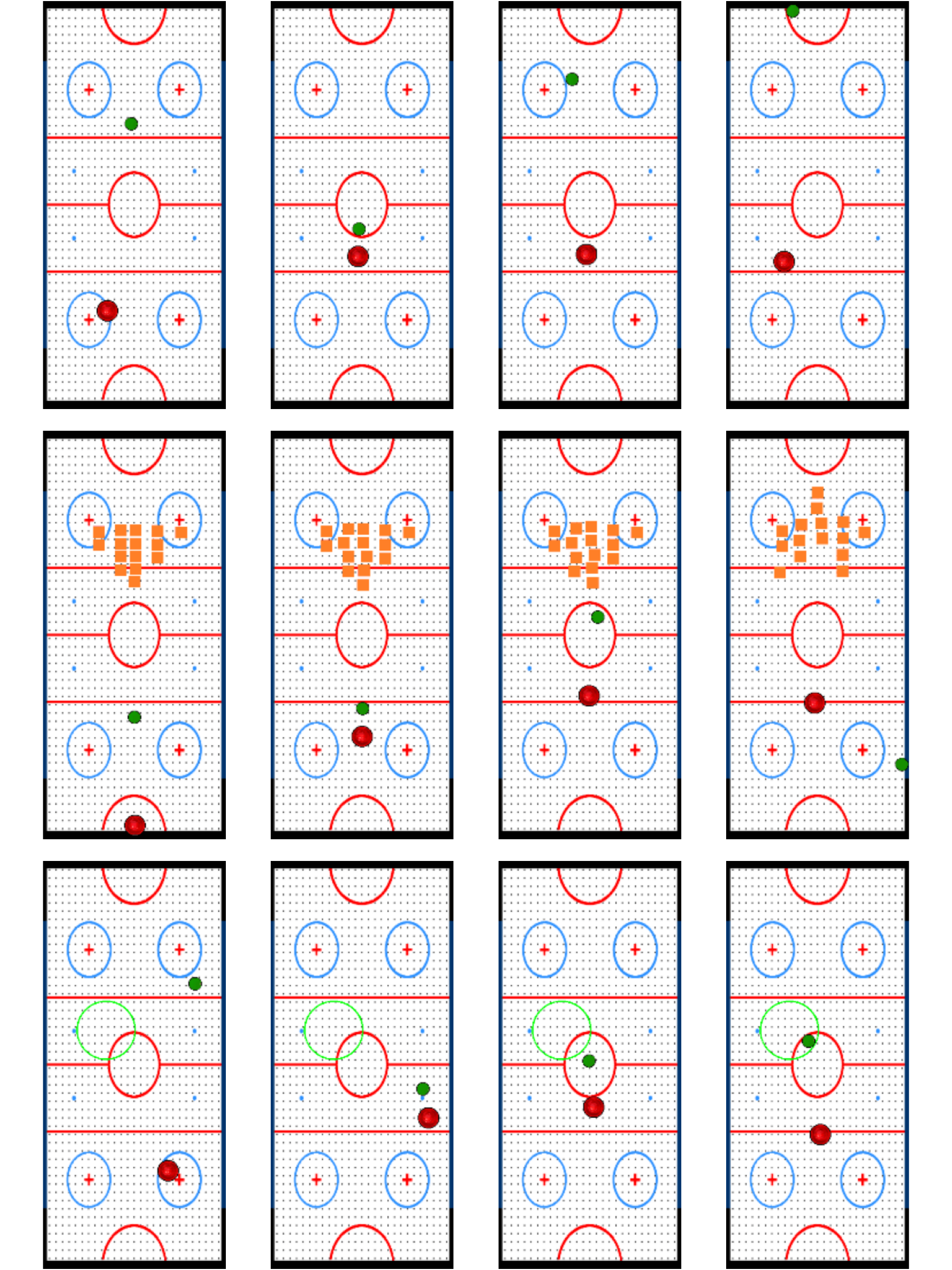}
    \caption{Execution rollouts in Box2D for various tasks. For each first frame the motion of the puck is downwards. Top row: The task where the policy tries to hit the puck to reach a minimum amount of upward velocity. Middle row: The task where the policy hits the puck into a crowd of blocks, causes them to spread. Bottom row: The task where the policy moves a puck into a goal region, shown as a green circle.}
    \label{fig:rollouts_Box2D}
\end{figure*}

\begin{figure*}[ht]
    \centering
    \centering
    \includegraphics[width=0.8\linewidth]{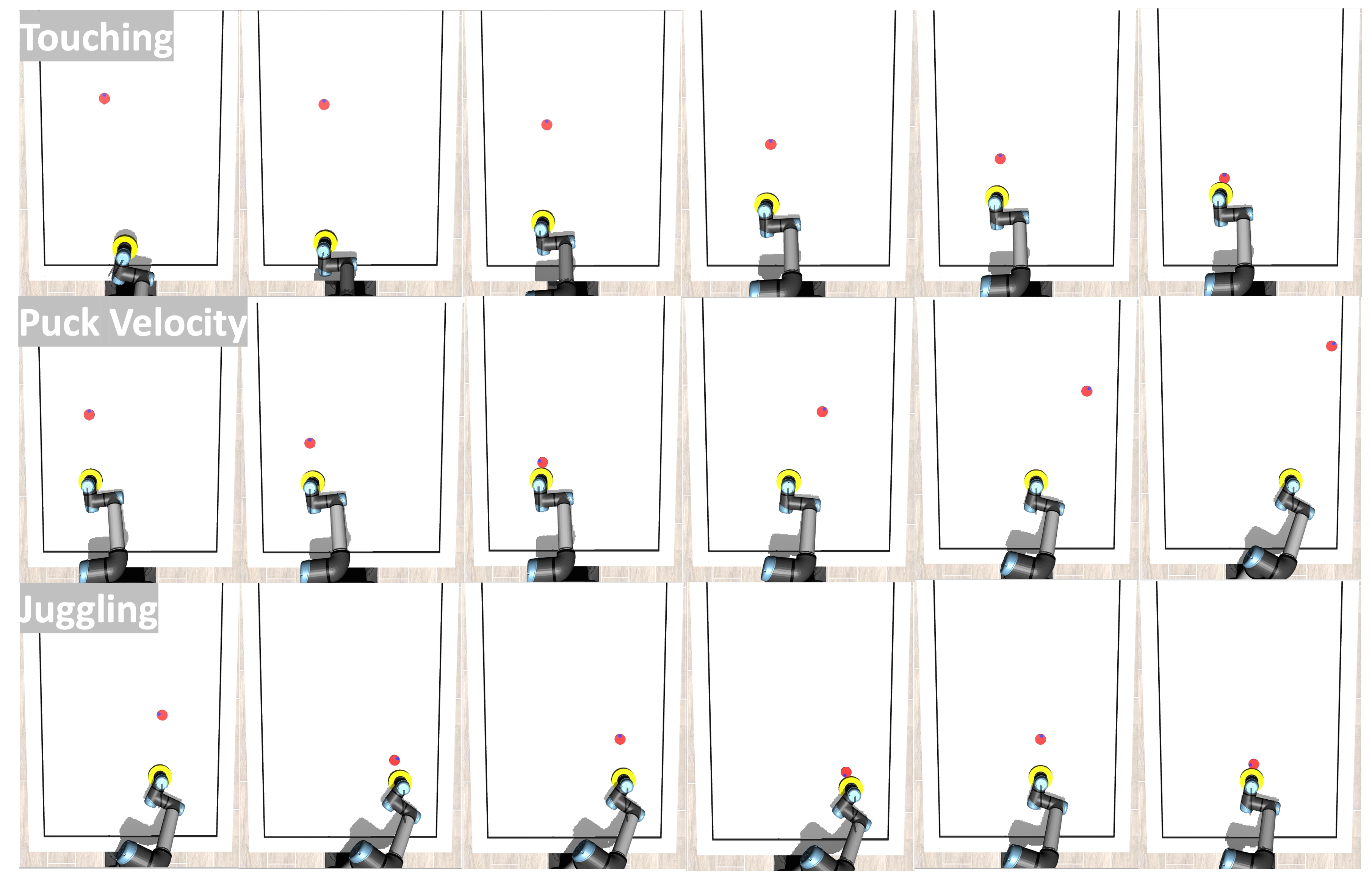}
    \caption{Execution rollouts in Robosuite simulator. The puck is circled in red for emphasis. Policies are trying to touch the puck, hit the puck with a minimum upward velocity, and juggle the puck.}
    \label{fig:rollouts_robosuite}
\end{figure*}

In this section, we briefly describe the  results of running various learning algorithms in the three domains. 
Overall, online RL performs the best among the baselines in simulation, showing that online interactions are crucial for solving our dynamic tasks. In the real world, all of our baselines fall short to human performance, leaving room for potential improvements for future work. To assess qualitative results, we include several figures in the Appendix illustrating the robot behavior in Box2D (Figure~\ref{fig:rollouts_Box2D}), Robosuite (Figure~\ref{fig:rollouts_robosuite}) and the real world (Figure~\ref{fig:rollouts_real}). We also discuss training curves in Appendix~\ref{sec:training_curves}.

\subsection{Behavior Cloning Results}
In the real world, using the dataset of approximately 128,000 frames of human-generated striking behavior gathered through two teleoperation systems, we train a network that maps puck and proprioceptive state information to actions. Because humans can struggle to hit using teleoperation, the learned models likewise struggle, while also being hampered by distribution mismatch. In the real world, we also experimented with an image-based policy using a ResNet-18  In the simulated domains, we behavior clone using 1M time steps from high-performing policies: if the RL policy performs well on a task, we behavior clone these policies for that task. If it does not, we use hindsight relabelling to retarget goals from a different task that does have high performance and use that. 

\subsection{Vanilla RL Results}
We assess Vanilla RL in simulation providing sufficient reward shaping to ensure that the agents can perform well, and find good performance in both simulators, as seen in Table~\ref{tab:exp-main}. In this table, we also describe the specific RL algorithms we employ. This provides evidence to validate the hypothesis that RL is an ideal choice in dynamic, interactive tasks where behavior cloning might struggle to utilize the nuanced distinctions in capabilities. In the real world, however, the random exploration necessary for RL from scratch prevents the robot from gathering enough meaningful feedback without wearing down the robot. 

\subsection{Offline RL Results}
Offline RL offers a mechanism for utilizing data collected offline from the agent, such as our teleoperation dataset, to learn a policy. For offline RL, we use IQL \cite{kostrikov2021iql}. Because Offline RL does not generate its own environment interactions, though, it can struggle more with distribution mismatch. However, this does allow us to assess Offline RL in the real world, where reward in dynamic, complex tasks outperforms behavior cloning.

\section{Conclusion}
We introduce an air-hockey-based evaluation domain for RL to evaluate RL in dynamic interactive environments. We provide the initial set of evaluations that support the hypothesis that in these kinds of domains, RL can outperform imitation learning in a real-world robotics setting. Most importantly, the suite of tasks, simulators, and collected human data opens the door for assessing a wide variety of RL settings from goal-conditioned RL to unsupervised skill learning to sim-to-real transfer, which we explore more in Appendix~\ref{sec:future_work}. Videos describing the project, of learned policies and human data can be found at \href{http://RLAirHockey.github.io}{RLAirHockey.github.io}.






\onecolumn
\twocolumn
\section{Appendix}
\setcounter{subsection}{0}
\subsection{Future Work And Relation with Manipulation Skills}
\label{sec:future_work}
\subsubsection{Roadmap for Manipulation Skills}
Robot manipulation has made tremendous progress in recent years. By creating Robot Air Hockey, we aim to push the robot manipulation research frontier in two major ways.

First, we explore the setting of dynamic object manipulation, which is understudied in the real world relative to quasistatic object manipulation because of many challenges. Compared to other dynamic robotics tasks (e.g. learning quadrupedal policies), it can be difficult to collect data for dynamic manipulation. Mainly, the low latency requirements and frequent (often human) resets required in these tasks bottleneck data’s quality and quantity. We hope that our system serves as a potential testbed and provides insights for future works where we address some of these challenges. Some of the insights we observed from creating a dynamic manipulation system are: 1) maintaining low latency is challenging, and 2) playing at a relatively high frame rate---20Hz or 50ms, even humans can struggle. We hope to follow up with experiments investigating human capabilities across different modalities of teleoperation. Furthermore, one of the most significant hurdles is the robot emergency-stopping, either because a learned policy takes cyclic actions that result in damaging resonant behavior, or because the robot jerks too quickly when changing direction, making the robot exceed its acceleration limits. Further investigation into action smoothing could provide higher-quality human demonstration data and offline policies.

Furthermore, recent work in robot learning has mostly focused on learning from demonstrations, where the data provided are usually optimal trajectories from humans or internet-scale data used to train representations. This optimality assumption is safe in quasistatic or low interaction environments, but in dynamic object manipulation, human teleoperators can lack the skill necessary to provide high-quality demonstrations. Robot air hockey considers one type of data that is relatively understudied: in-domain low-reward interaction data. Currently, we have collected robot data via teleoperation, and we are also planning to collect robot interaction data autonomously soon. As a result, this will open up avenues for many algorithms that are not limited to learning from demonstrations. For instance, our system allows us to assess RL algorithms and offers the opportunity to assess many forms of RL, such as goal-conditioned, offline or sim-to-real methods, to name a few. Because of the suboptimality of demonstration data, RL is an ideal tool for this setting, as suggested by the results in this work where offline RL outperforms other learning methods such as behavior cloning. Since related work assessing RL directly on a physical robot, especially in a dynamic manipulation setting, is limited, this work offers offline RL assessment in a dynamic, interactive real-world set of tasks.

Our roadmap for future robot manipulation research is to learn from suboptimal interaction data and solve dynamic manipulation tasks in the real world. Current approaches have achieved incredible performance in quasistatic manipulation tasks, thanks to their abilities to harness optimal demonstration data and understand the task dynamics. However, doing so for dynamic manipulation is much more difficult. Although the scale of large models and datasets increases, in-domain data for dynamic tasks is still hard to collect. Even if one can collect in-domain data for these tasks, collecting optimal data is difficult, let alone deriving an accurate dynamics model for the task. Therefore, we believe this system is a powerful potential testbed for applying RL in the real world where we envision achieving super-human level performance on challenging, dynamic manipulation tasks.
\subsubsection{Future Work}
Perhaps the most obvious setting for RL in air hockey is \textbf{goal-conditioned RL}. The inverted table is inherently goal-conditioned in the sense of striking the puck to a desired position, and one of the most challenging of our existing tasks is striking the puck to a desired position \textit{with} a desired velocity. However, many more goal-conditioned settings exist, including using the puck to hit an object to a location, achieving a sequence of goals for the end effector, or getting the puck into a desired goal state relative to other objects.

Another clear setting for future RL assessment of the robot air hockey testbed, considering the multiple simulation environments of increasing fidelity, is to utilize offline data generated by a higher fidelity simulator to train policies using another simulator \cite{hanna2021grounded}. \textbf{Sim-to-real transfer} can be iterated on quickly through 2D-to-3D sim-to-sim transfer and then tested on 3D-to-real transfer. Similarly, \textbf{model-based RL} can utilize the simpler simulated environment models and transfer to model more complex real-world dynamics, especially using the offline data.

Another consequence of the paired simulators is to provide a natural curriculum for \textbf{curriculum learning}. Within single tasks, depending on the initialization of target objects tasks can range from easy to hard, as well as between simulators.  Also, because many of the tasks are derivative of each other, this leaves the room open for \textbf{transfer learning or meta-learning}, where learning on some tasks will benefit downstream learning. As an extension, \textbf{unsupervised learning or skill learning} make a lot of sense in many of the environments since certain strikes and hits will be useful across multiple tasks. While hard-coding these behaviors would be challenging, learning them from the offline dataset or from experience could allow agents to perform complex behaviors like hitting a target block to a desired location.

The air hockey setting itself offers possibilities for object-based factorization. While this work only provided a few additional objects beyond the puck and paddle, a wider variety of objects are actively being incorporated to investigate \textbf{object-based generalization} algorithms. For example, adding in target objects to push, obstacles for the puck, negative or positive reward regions, or even regions of different physics are all possible.  Furthermore, by modifying real-world parameters such as table angle or paddle shape, we can assess causal or model-based RL. Finally, especially as more objects are added, visual complexity opens the door for more complex object detection than simple color segmentation.

With many manipulation tasks focusing on tasks that can often be performed with high precision for human demonstrators, such as pick-and-place, this system offers the opportunity for \textbf{superhuman performance}, especially on even more challenging tasks, such as two-puck juggling. RL is an ideal tool for this setting, as suggested by the results in this work, but RL directly on the robot is limited. 

While the existing system shows a wide range of capabilities, the real-world system illuminates several insights about running learned policies in a \textbf{high-speed setting}. In particular, maintaining high latency is challenging, and we found that even humans playing at 20Hz can struggle. We hope to follow up with experiments investigating human capabilities in teleoperation. Furthermore, one of the most significant hurdles is the robot emergency-stopping, either because a learned policy takes cyclic actions that result in damaging resonant behavior, or because the robot jerks too quickly when changing direction. Further investigation into action smoothing could provide higher-quality human demonstration data and offline policies.

As an immediate direction, the air hockey testbed offers \textbf{offline RL} assessment in a dynamic, interactive real-world set of tasks. The initial results suggest that RL is preferable because of the difficulty of gathering high-quality demonstrations, and further investigation into offline and mixed offline and online methods can provide substantial benefits.

We are looking into \textbf{multi-agent settings}. Air hockey is itself an adversarial game. Beyond this, simulators can be modified to include collaborative settings such as rallying the puck between multiple paddles or multiple puck juggling with multiple paddles. Furthermore, we can extend the adversarial setting to more than two goal regions and opposing paddles in a free for all. On the physical robot, by incorporating an additional UR5 at the opposite end of the table, we can realize some of these settings. Alternatively, while humans far outstrip the physical capabilities of the UR5, adversarial play can be between two humans playing against each other through the robot. 

Finally, the teleoperation systems offer several means of assessing \textbf{human-robot interaction}. Some possible directions for this include simple assessments of human skill levels over play on the robot or frustrations with robot capabilities, to more complex assessments of humans being able to detect or react to another human playing, or a learned policy. 

In summation, we believe that the robot air hockey testbed offers ample opportunities to assess a significant range of RL capabilities across multiple domains. While there are certainly limitations, a shared interface can facilitate proof of concept work. 

\subsection{Training Curves}
\label{sec:training_curves}
\begin{figure*}[ht]
    \centering
    \includegraphics[width=0.75\textwidth]{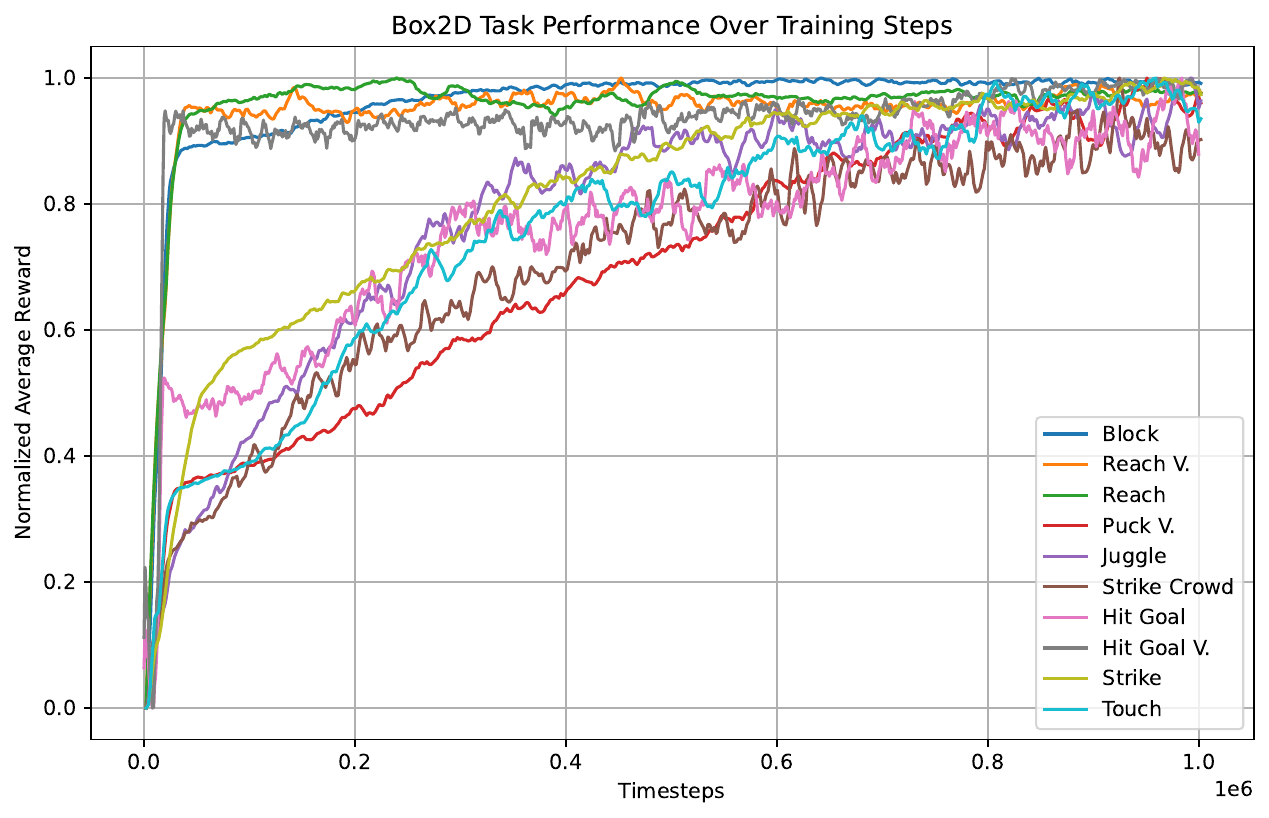}
    \caption{Training curve for all tasks in Box2D. For vanilla RL tasks, the rewards are averaged across 5 seeds, while for goal-conditioned RL we use 1 seed. Rewards are then normalized to to the [0, 1] range with respect to the minimum and maximum reward seen for each task. Tasks in which performance converges quickly indicate that the task is either trivially easy (reaching a position with the paddle, reaching a position with the paddle with a desired velocity) or too difficult (moving a block, hitting a puck into a goal position with desired velocity).}
    \label{fig:training_Box2D}
\end{figure*}

\begin{figure*}[t]
    \centering
    \begin{subfigure}[t]{0.45\linewidth}
    \centering
    \includegraphics[width=\linewidth]{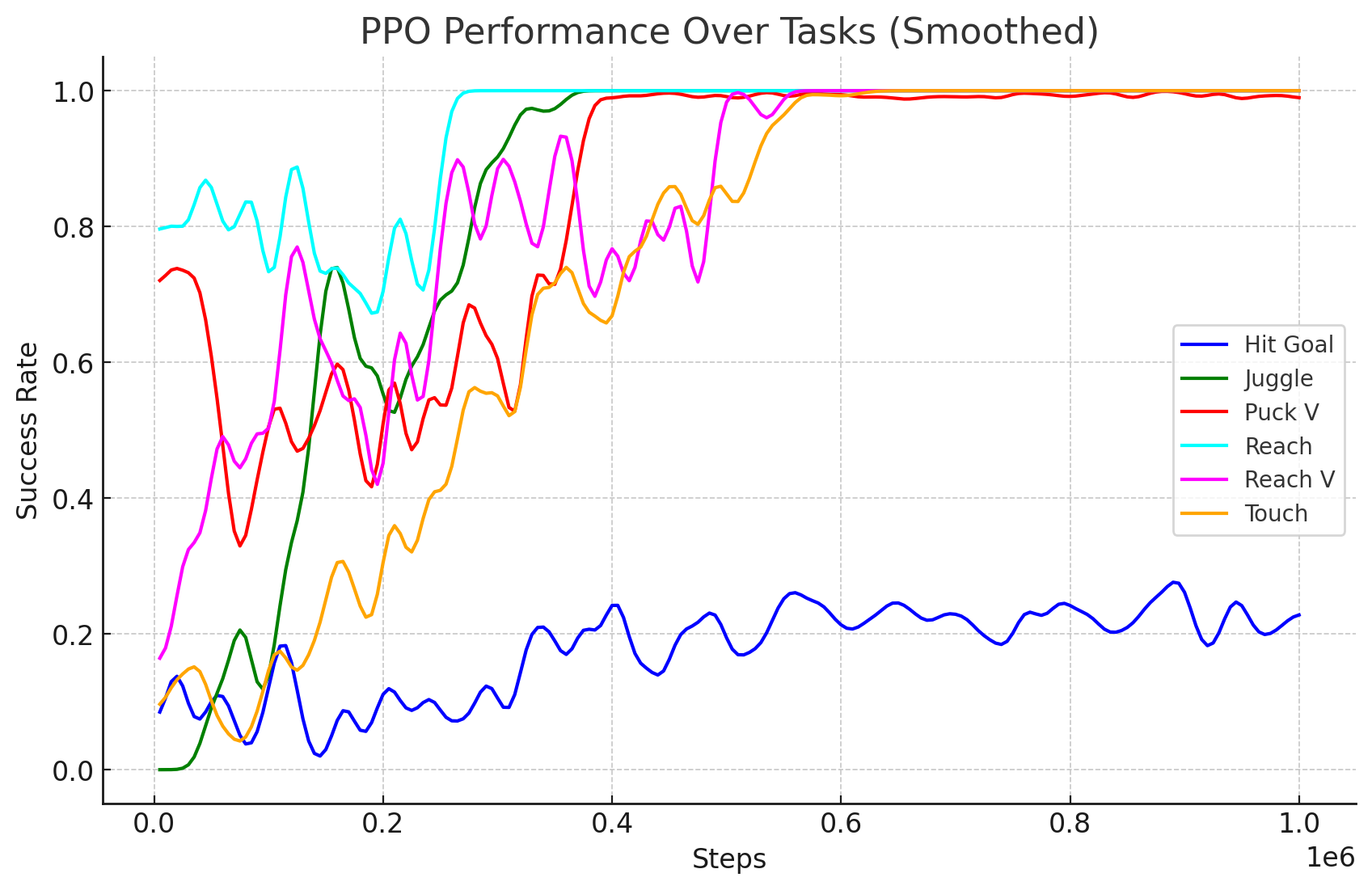}
    \caption{}
    \end{subfigure}
    ~
    \begin{subfigure}[t]{0.45\linewidth}
    \centering
    \includegraphics[width=\linewidth]{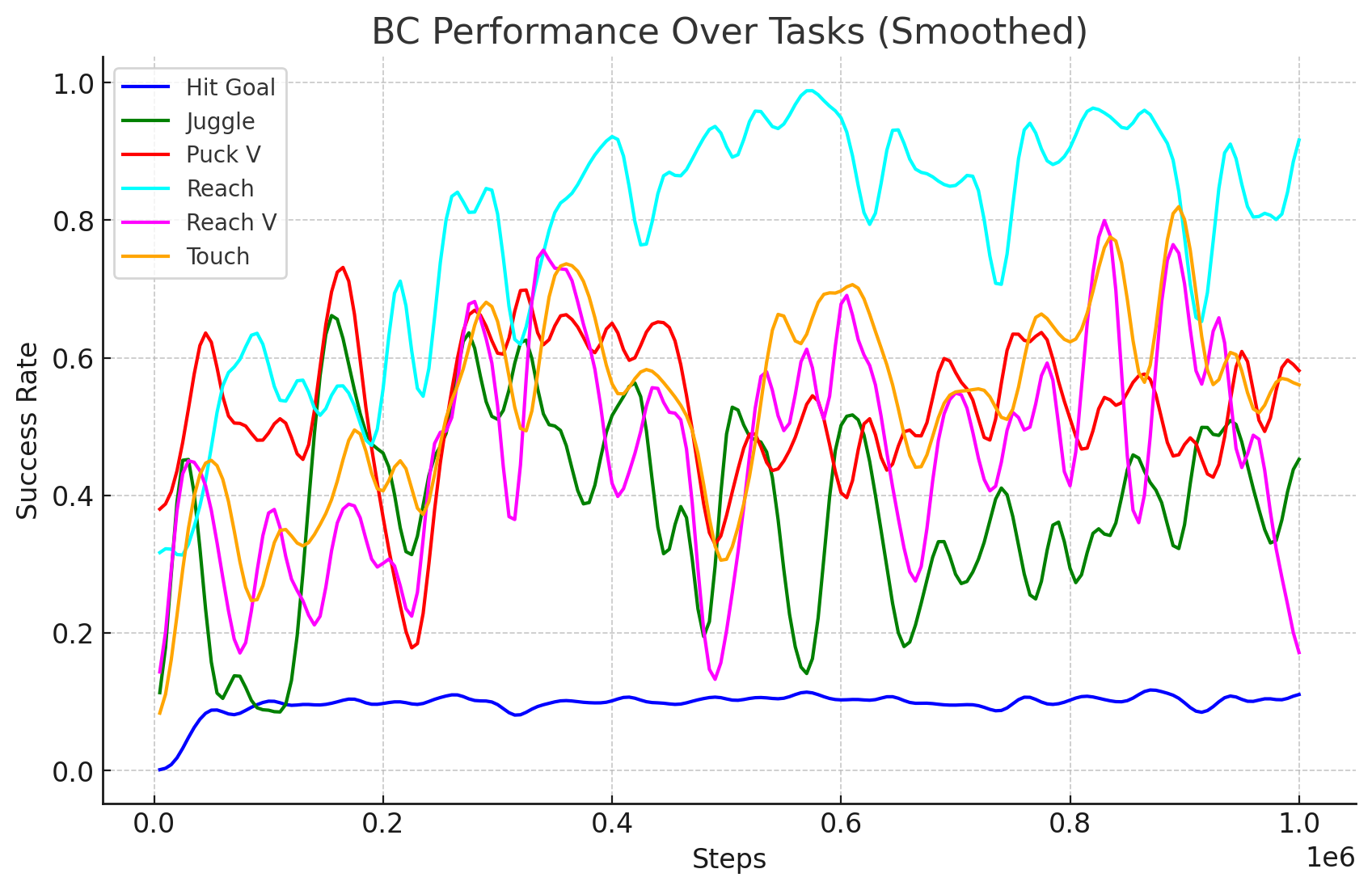}
    \caption{}
    \end{subfigure}
    \begin{subfigure}[t]{0.45\linewidth}
    \centering
    \includegraphics[width=\linewidth]{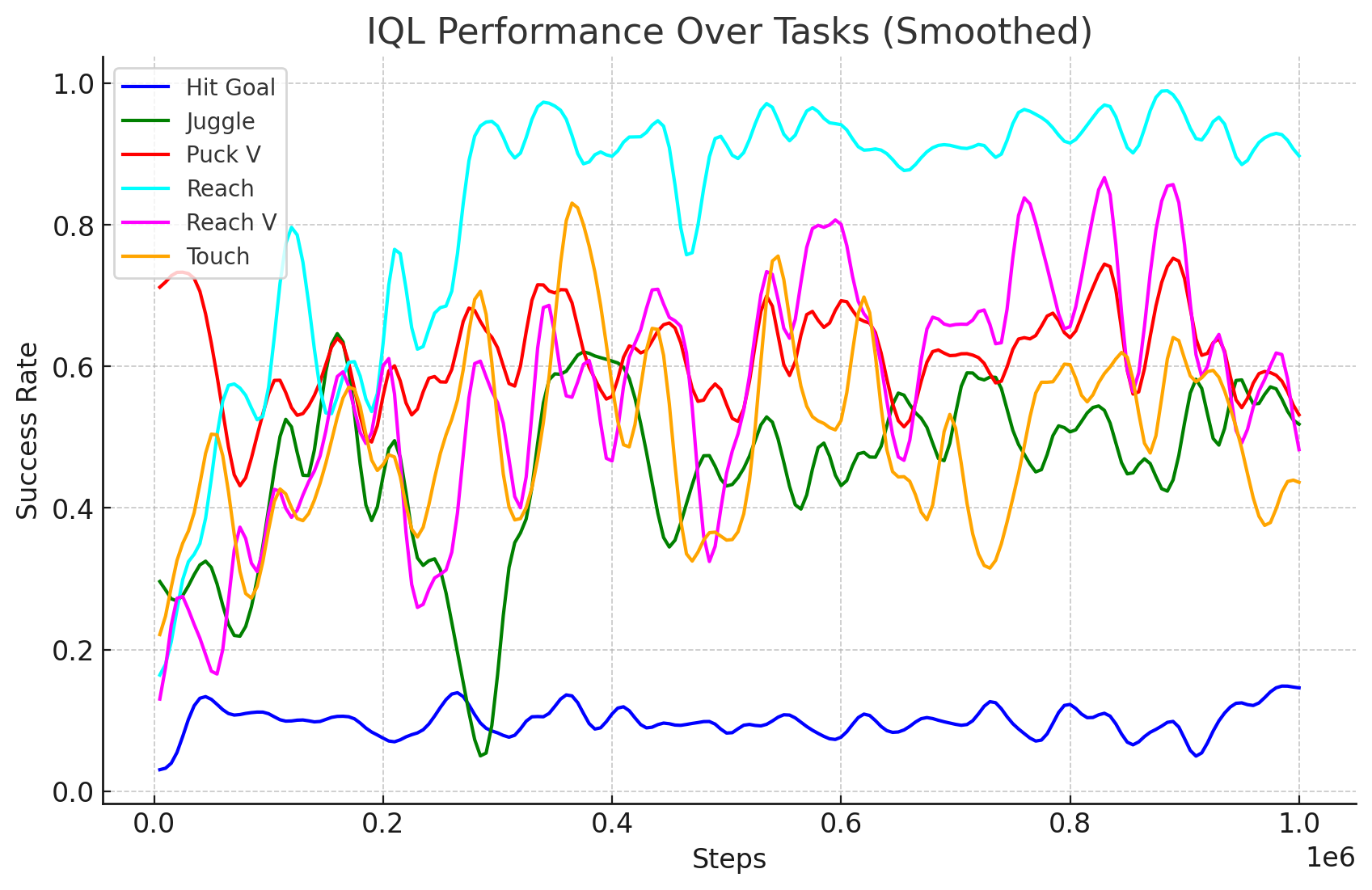}
    \caption{}
    \end{subfigure}
    \caption{Robosuite Training curves. \textbf{(a)}: Vanilla RL with PPO, following the implementation of CleanRL library \cite{huang2022cleanrl}. \textbf{(b)} Behavior Cloning with data collected the "expert" policy (a trained PPO policy). \textbf{(c)} IQL \cite{kostrikov2021iql} using the same offline data, using asymmetric $\tau$ set to 0.6.} 
    \label{fig:robosuite_training}
\end{figure*}

\begin{figure*}[ht]
    \centering
    \includegraphics[width=1.0\textwidth]{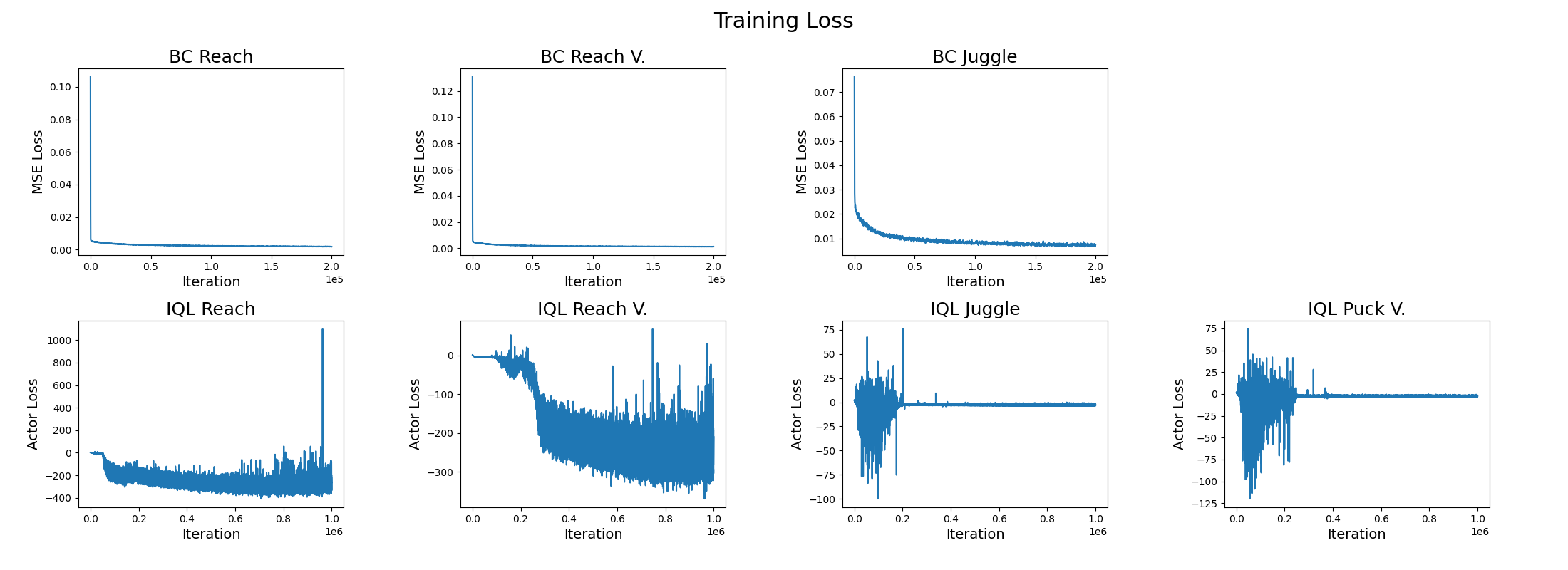}
    \caption{Loss (MSE of behavior cloning and Actor Loss for IQL) curves for training using data collected from the real robot. Since humans did not distinguish puck velocity and juggling, we only trained a single behavior cloning policy (BC Juggle).}
    \label{fig:training_real}
\end{figure*}

We illustrate the training curves as normalized reward over the number of timesteps for Box2D (Figure~\ref{fig:training_Box2D}), success rate over 15 episodes over the number of training timesteps for Robosuite (Figure~\ref{fig:robosuite_training}), mean squared error for behavior cloning on the real robot, and actor loss for IQL over number of iterations of training (Figure~\ref{fig:training_real}). The real robot uses different losses because we do not evaluate during training, as that would require loading partial networks onto the real robot, which would greatly extend training time.

\subsection{Puck Hitting dataset}
\label{sec:puck_detection}
This section will describe how we extract the state of the puck on the table from the collected videos and show exemplary trajectories from our dataset. We collected videos of several participants attempting to hit the puck after being dropped from the far end of the table using the various teleoperation modalities.

\begin{figure*}[t!]
    \centering
    \begin{subfigure}[t]{0.22\linewidth}
    \centering
    \includegraphics[width=\linewidth]{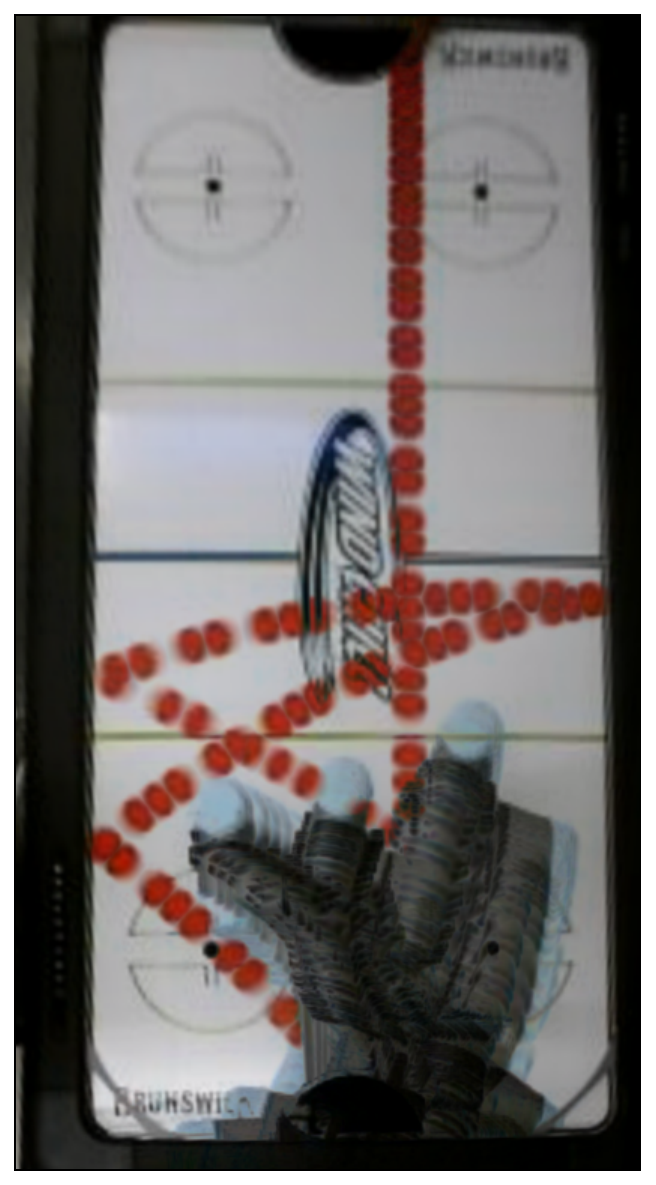}
    \caption{Stacked images representing the video of a trajectory collected with \textbf{Shadow-teleop}}
    \end{subfigure}
    ~
    \begin{subfigure}[t]{0.22\linewidth}
    \centering
    \includegraphics[width=\linewidth]{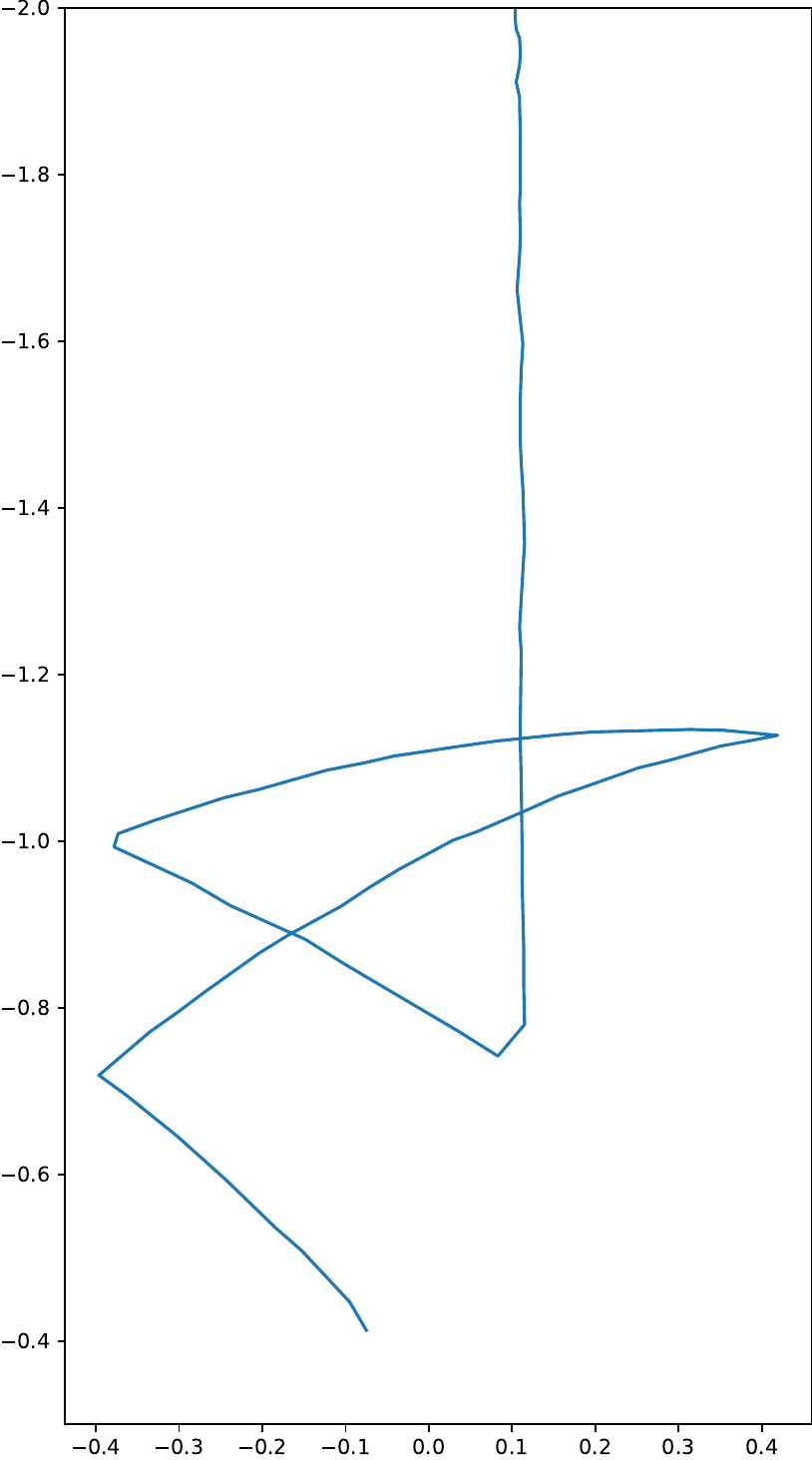}
    \caption{Extracted state trajectory of puck data collected with \textbf{Shadow-teleop}}
    \end{subfigure}
    ~   
    \begin{subfigure}[t]{0.22\linewidth}
    \centering
    \includegraphics[width=\linewidth]{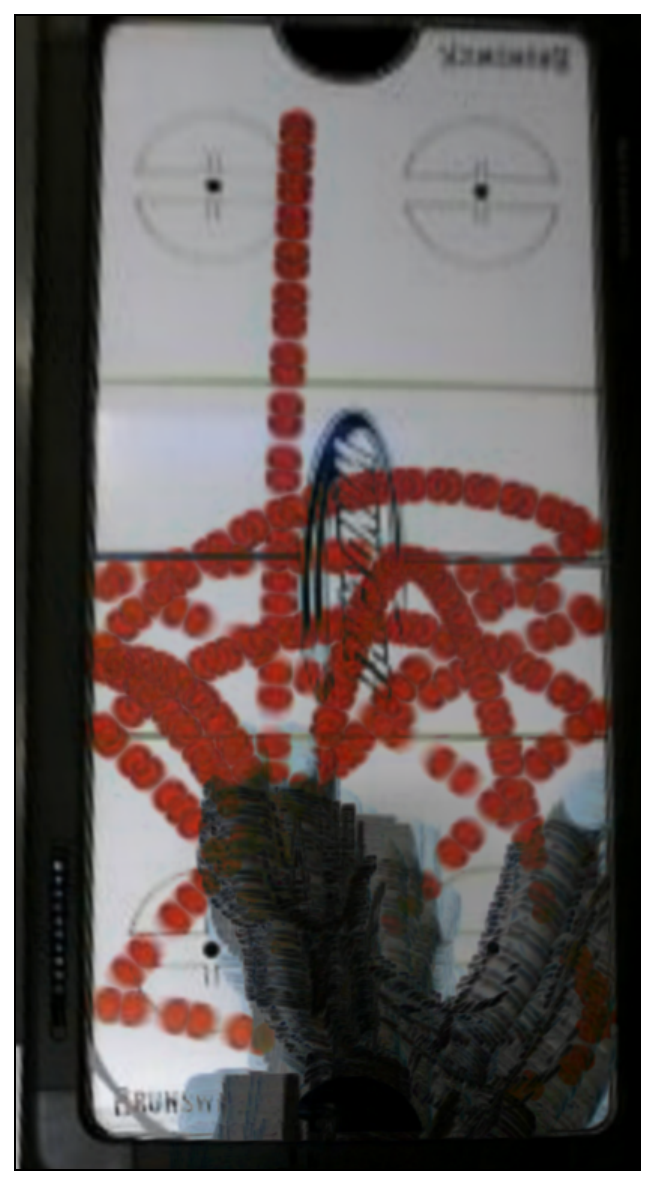}
    \caption{Stacked images representing the video of a trajectory collected with \textbf{Mouse-teleop}}
    \end{subfigure}
    ~
    \begin{subfigure}[t]{0.22\linewidth}
    \centering
    \includegraphics[width=\linewidth]{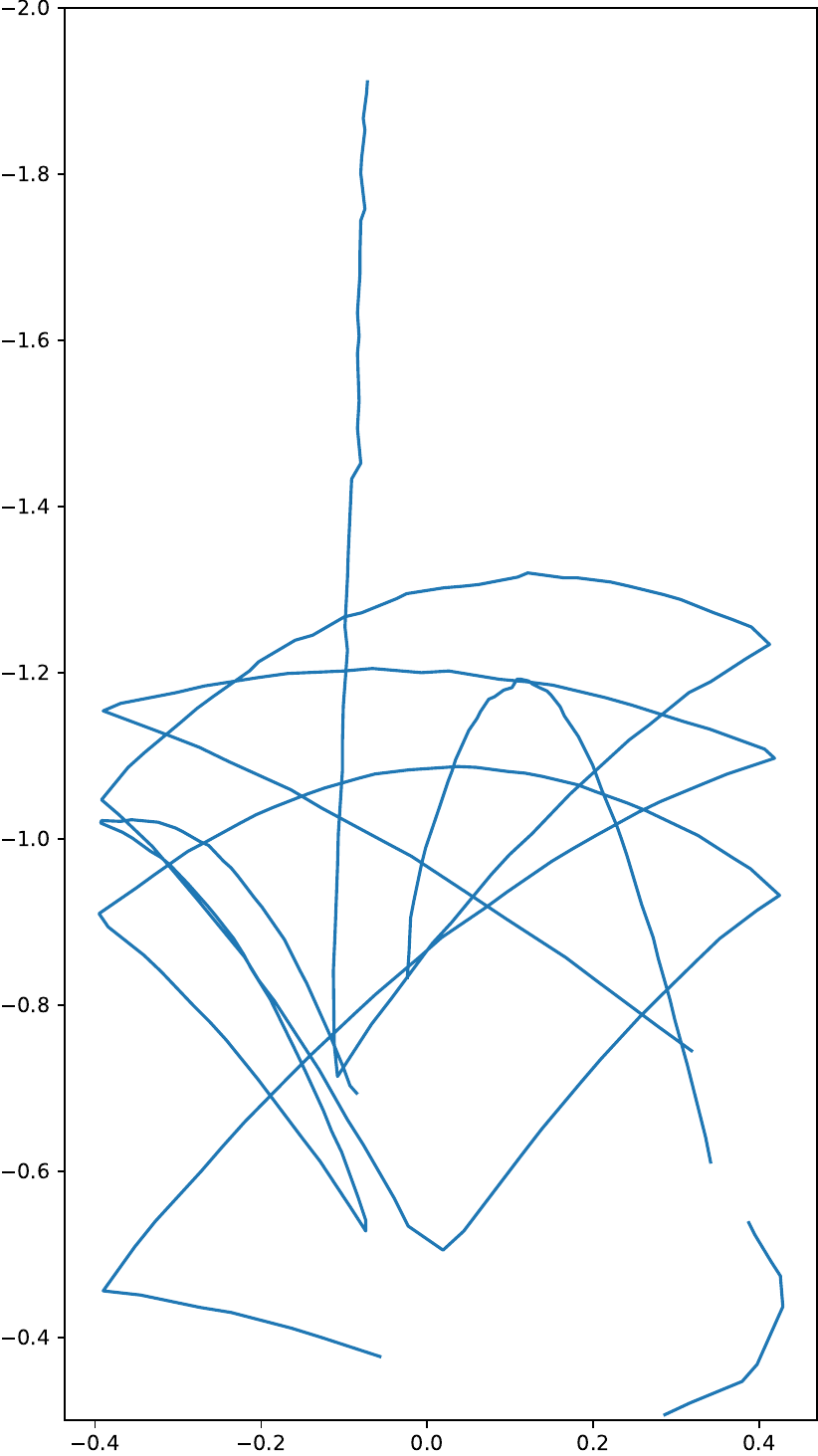}
    \caption{Extracted state trajectory of puck data collected with \textbf{Mouse-teleop}}
    \end{subfigure}
    
    \caption{Puck-hitting trajectories collected with teleoperation modalities. Generally, mouse-teleop is more responsive, and thus participants could strike the puck more easily. Nonetheless, participant skill was the primary factor when assessing the quality of a gathered trajectory.}
    \label{fig:puck_detect}
\end{figure*}

\textbf{Puck state extraction.} We use red or green pucks because they are the easiest to locate on the air hockey table using HSV color segmentation. To find the location of the puck on the air hockey table, we first apply a homography transformation to the image to account for the distortions introduced by the camera. We then apply a mask to each frame that indicates whether a pixel falls within the color bounds for the red or green puck. Lastly, we find the median pixel location of all the masked pixels and apply an affine transform that maps from the pixel location in the image to the robot base's reference frame. 
\textbf{Puck trajectories}
In Figure \ref{fig:puck_detect}, we illustrate the extracted trajectories from shadow-teleop and mouse-teleop. Gaps in the puck's trajectory indicate where the puck was not detected, for example, if it was occluded by the robot.

\bibliographystyle{IEEEtran} 
\bibliography{bibliography} 

\end{document}